\documentclass[runningheads]{llncs}
\usepackage{graphicx}
\usepackage{comment}
\usepackage{amsmath,amssymb}
\usepackage{color}
\usepackage{bm}
\usepackage{booktabs}
\usepackage{appendix}

\graphicspath{{figs/}}

\makeatletter
\newcommand{\printfnsymbol}[1]{%
  \textsuperscript{\@fnsymbol{#1}}%
}
\makeatother

\begin{document}
\pagestyle{headings}
\mainmatter
\def\ECCVSubNumber{1248}

    \title{Fast Video Object Segmentation using the Global Context Module}
    \titlerunning{Fast Video Object Segmentation using the Global Context Module}
    \author{
        Yu Li\inst{1}\thanks{Both authors contributed equally. This work was done during Zhuoran's internship at Tencent.} \and
        Zhuoran Shen\inst{2}\printfnsymbol{1}\and
        Ying Shan\inst{1}
    }
    \authorrunning{Y. Li et al.}
    \institute{
        Applied Research Center (ARC), Tencent PCG
        \and
        The University of Hong Kong
    }

\maketitle

\begin{abstract}
We developed a real-time, high-quality semi-supervised video object segmentation algorithm. Its accuracy is on par with the most accurate, time-consuming online-learning model, while its speed is similar to the fastest template-matching method with sub-optimal accuracy. The core component of the model is a novel global context module that effectively summarizes and propagates information through the entire video. Compared to previous approaches that only use one frame or a few frames to guide the segmentation of the current frame, the global context module uses all past frames. Unlike the previous state-of-the-art space-time memory network that caches a memory at each spatio-temporal position, the global context module uses a fixed-size feature representation. Therefore, it uses constant memory regardless of the video length and costs substantially less memory and computation. With the novel module, our model achieves top performance on standard benchmarks at a real-time speed.
\keywords{video object segmentation, global context module}
\end{abstract}

\section{Introduction}

Video object segmentation~\cite{snapcut,cut-and-paste,live-cut,interact-cutout} aims to segment a foreground object from the background on all frames in a video. The task has numerous applications in computer vision. An important one is intelligent video editing. As videos become the most popular form of media on mass content platforms, video content creation is getting increasing levels of attention. Object segmentation on each frame with image segmentation tools is time-consuming and has poor temporal consistency. Semi-supervised video object segmentation tries to solve the problem by segmenting the object in the whole video given only a fine object mask on the first frame. This problem is challenging since object appearance might vary drastically over time in a video due to pose changes, motion, and occlusions, \textit{etc}.

\begin{figure}
    \includegraphics[width=\columnwidth]{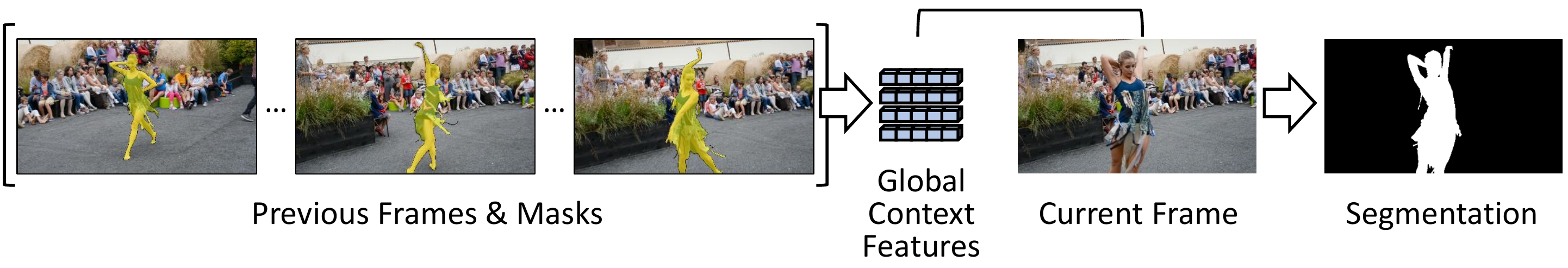}
    \caption{Our video object segmentation method creates and maintains fixed-size global context features for previous frames and their object masks. These global context features can guide the segmentation on the incoming frame. The global context module is efficient in both memory and computation and achieves high segmentation accuracy. See more details below.}
    \label{fig:teaser}
\end{figure}

With the success in many computer vision tasks, deep learning techniques are widely used in video object segmentation recently. The essence is to learn an invariant representation that accounts for the variation of object appearance across frames. Some early works~\cite{osvos,osvoss,masktrack,onavos} in semi-supervised video object segmentation use the first frame to train a model using various data augmentation strategies, which are commonly referred to as online learning-based methods. These methods usually obtain accurate segmentation that are robust to occlusions. However, online learning incurs huge computational costs that lead to several seconds of processing per frame. Another direction is propagation-based methods, \textit{e.g.}~\cite{fvos}, which rely on the segmentation of the previous frame to infer for the current frame. These methods are simple and fast, but usually have sub-optimal segmentation accuracies. These methods cannot handle occlusions and may suffer from error drifts. Some later works take the advantages of both directions, use both the first frame and the previous frame~\cite{feelvos,ranet,rgmp,osmn}, and achieve both high accuracy and fast processing speed.

A recent work~\cite{stm} makes a further step to use all previous frames with the corresponding object segmentation results to infer the object mask on the current frame. It proposes a novel space-time memory (STM) module that stores the segmentation information at each processed frame, \textit{i.e.}, this memory module saves information at all spatio-temporal locations on all previous frames. When working on a new frame, a read operation is used to retrieve relevant information from the memory by performing a dense feature correlation operation in both temporal and spatial dimensions. By using this memory that saves all guidance information, the method is robust against drastic object appearance changes and occlusions. It produces promising results and achieves state-of-the-art performance on multiple benchmark datasets.

While STM achieved state-of-the-art performance by making full use of the information from previous frames, leveraging the space-time memory is costly, especially on long videos. As the space-time memory module keeps creating new memories to save new information and put it together with old memories, the computational cost and memory usage in the feature correlation step increase linearly with the number of frames. This makes the method slower and slower while processing and may easily cause GPU memory overflow. To resolve this issue, the authors propose to reduce the memory saving frequency, \textit{e.g.} saving a new memory every 5 frames. However, the linear complexity with time still exists, and such reduction in memorization frequency defeats the original purpose to utilize information from every previous frame.

In this paper, building upon the idea of STM to use information in all past frames, we develop a compact global representation that summarizes the object segmentation information and guides the segmentation of the next frame. This representation automatically updates when the system moves forward by a frame. The core component of it is a novel global context module (illustrated in Fig.~\ref{fig:teaser}). By keeping only a fixed-size set of features, our memory and computational complexities for inference are light and do not increase with time, in comparison to the linear complexities of the STM module. We show that using this highly efficient global context module, our method is about three times faster than STM and do not need to worry about the memory usage. The performance of our method on the single object segmentation benchmark DAVIS 2016 in terms of segmentation accuracy is on par with the state-of-the-art. The results of our method on more challenging multiple object segmentation benchmarks DAVIS 2017 and YouTube-VOS are highly competitive.

The contribution of our paper can be summarized as:

\begin{itemize}
\setlength{\itemsep}{0pt}
    \item We propose a novel global context module that reliably maintains segmentation information of the entire video to guide the segmentation of incoming frames.
    \item We implement the global context module in a light-weight way that is efficient in both memory usage and computational cost.
    \item Experiments on DAVIS and YouTube-VOS benchmarks show that our proposed global context module can achieve top accuracy in video object segmentation and is highly efficient that runs in real time.
\end{itemize}

\section{Related Works}
\label{sec:review}

\subsubsection{Online-learning methods}
Online-learning methods usually fine-tune a general object segmentation network on the object mask of the start frame to teach
the network to identify the appearance of the target object in the remaining video frames~\cite{osvos}. They use online adaptation~\cite{onavos}, instance segmentation information~\cite{osvoss}, data augmentation techniques~\cite{lucid}, or an integration of multiple techniques~\cite{premvos}. Some methods report that online learning boosts the performance of their model~\cite{dyenet,ranet}. While online learning can achieve high-quality segmentation and is robust against occlusions, it is computationally expensive as it requires fine-tuning for each video. This huge computational burden makes it impractical for most applications.

\subsubsection{Offline-learning methods}
Offline-learning methods use strategies like mask propagation, feature matching, tracking, or a hybrid strategy. Propagation-based methods rely on the segmentation mask of the previous frame. It usually takes the previous mask as an input and learns a network to refine the mask to align it with the object in current frame. A representative work is MaskTrack~\cite{masktrack}. This strategy of using the previous frame is also used in ~\cite{cinm,osmn}. Many works~\cite{cheng2017segflow,jang2017online,masktrack} also use optical flow~\cite{flownet,flownet2} in the propagation process. Matching-based methods~\cite{pml,videomatch,shin2017pixel,feelvos,ranet,rgmp} treat the first frame (or intermediate frames) as a template and match the pixel-level feature embeddings in the new frame with the templates. A more common setup is to use both the first frame and the previous frame~\cite{feelvos,ranet,rgmp}, which covers both long-term and short-term object appearance information.

\subsubsection{Space-time memory network}
The STM network~\cite{stm} is a special feature matching-based method which performs dense feature matching across the entire spatio-temporal volume of the video. This is achieved by a novel space-time memory mechanism that stores the features at the each spatio-temporal location. The space-time memory module mainly contains two components and two operations. The two components are a key map and a value map where the keys encode the visual semantic embeddings for robust matching against appearance variations, and the values store detailed features for guiding the segmentation. The memory write operation simply concatenates the key and value maps generated on past frames and their object segmentation masks. When processing a new frame, the memory read operation uses the keys to match and find the relevant locations in the spatio-temporal volume in the video. Then the features stored in the value maps at those locations are retrieved to predict the current object mask.

\begin{figure*}[tb]
    \includegraphics[width=\linewidth]{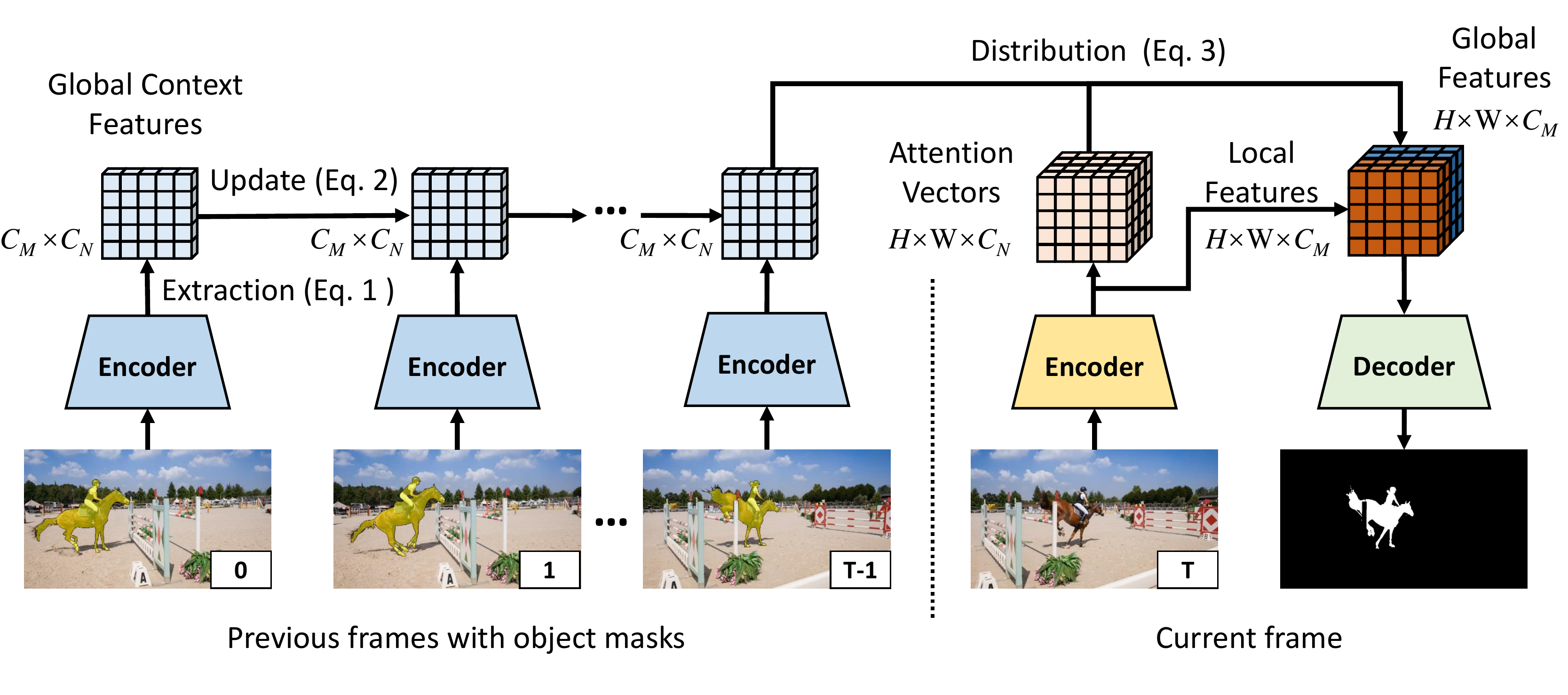}
    \caption{This is an overview of our pipeline. Our network encodes past frames and their masks to a fixed-size set of global context features. These vectors are updated by a simple rule when the system moves to the next frame. At the current frame, a set of attention vectors are generated from the encoder to retrieve relevant information in the global context to form the global features. Local features are also generated from the encoder output. The global and local features are then concatenated and passed to a decoder to produce the segmentation result for this frame. Note that there are two types of encoders, the blue one for past frames and masks (four input channels), and the orange one for the current frame (three input channels).}
    \label{fig:pipeline}
\end{figure*}

\section{Our Method}
\label{sec:method}
\subsection{Global Context Module}

The space-time memory network \cite{stm} achieves great success in video object segmentation with the space-time memory module. The STM module is an effective module that queries features from a set of spatio-temporal features encoding previous frames and their object masks to guide the processing of the current frame. In this way, the current frame is able to use the features from semantically related regions and the corresponding segmentation information in past frames to generate its features. However, the STM module has a drawback in efficiency in that it stores a pair of key and value vectors for each location of each frame in the memory. These feature vectors are simply concatenated over time when the system moves forward and their sizes keep increasing. This means its resource usage is highly sensitive to the spatial resolution and temporal length of the video. Consequently, the module is limited to have memories with low spatial resolutions, short temporal spans, or reduced memory saving frequency in practice. To remedy this drawback, we propose the global context (GC) module. Unlike the STM module, the global context module keeps only a small set of fixed-size global context features yet retains almost the same representational power compared to the STM module. Fig.~\ref{fig:pipeline} shows the overall pipeline of our proposed method using the global context module. The main structure is an encoder-decoder and the global context module is built on top of the encoder output, similarly to ~\cite{stm}.  There are mainly two operations in our pipeline, namely 1) context extraction and update on a processed frame and its object mask, and 2) context distribution to the current frame under processing. There are two types of encoders used which produce features of $ H \times W$ resolution and $C$ channels. One takes a color image and an object mask (ground truth mask for the start frame and segmentation results for intermediate frames) to encode the frame and segmentation information into the feature space. Another encoder encodes the current frame to a feature embedding. We distribute the global context features stored in the global context module and combine the distributed context features with local features. Then, a decoder is used to produce the final object segmentation mask on this frame.

\begin{figure*}[tb]
    \includegraphics[width=\linewidth]{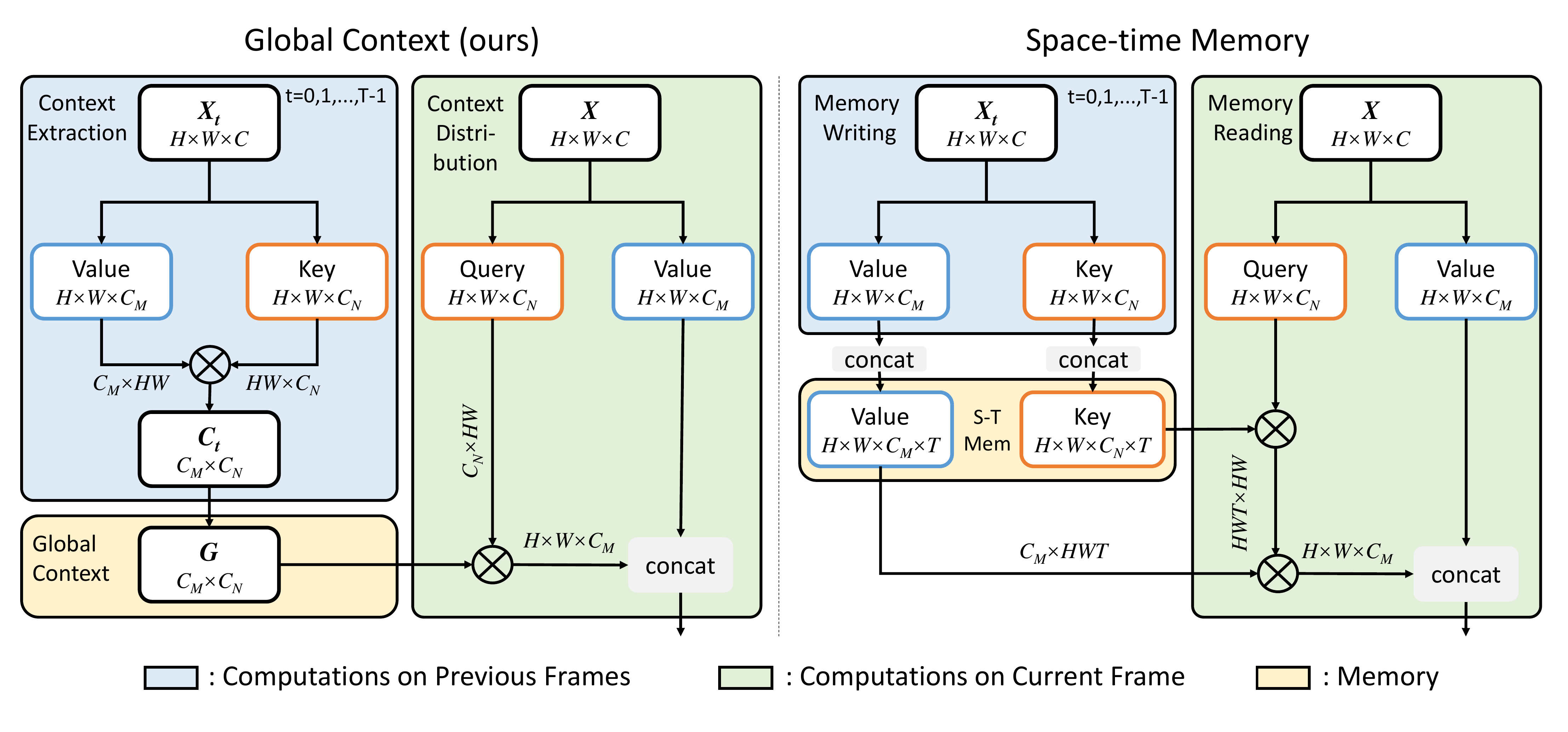}
    \caption{The detailed implementation of the global context module with a comparison to the space-time memory module in \cite{stm}.}
    \label{fig:compare}
\end{figure*}

\subsubsection{Context Extraction and Update}

When extracting the global context features, the global context module first generates the keys and the values, following the same procedure as in the STM module. The keys and the values have size $H \times W \times C_N$ and $H \times W \times C_M$ respectively, where $C_N$ and $C_M$ are the numbers of channels used. Unlike the STM module that directly stores the keys and values, the global context module puts the keys and the values through a further step called global summarization.

The global summarization step treats the keys not as $H \times W$ vectors of size $C_M$ each for a location, but as $C_M$ 1-channel feature maps each as an unnormalized weighting over all locations. For each such weighting, the global context module computes a weighted sum of the values where each value is weighted by the scalar at the corresponding location in the weighting. Each such weighted sum is called a global context vector. The global context module organizes all $C_M$ global context vectors into a global context matrix as the output of global summarization. This step can be efficiently implemented as a matrix product between the transpose of the key matrix and the value matrix. Following is an equation describing the context extraction process of the global context module:
\begin{equation}
\label{eq:eq1}
    \bm{C}_t = k(\bm{X}_t)^\mathsf{T}v(\bm{X}_t),
\end{equation}
where $t$ is the index of the current frame, $\bm{C}_t$ is the context matrix of this frame, $\bm{X}_t$ is the input to the module (output of the encoder), and $k, v$ are the functions that generate the keys and values. Having obtained $\bm{C}_t$, the rule for updating the global context feature is simply
\begin{equation}
\label{eq:eq2}
    \bm{G}_t = \frac{t-1}{t}\bm{G}_{t-1} + \frac{1}{t}\bm{C}_t,
\end{equation}
where $\bm{G}_t$ is the global context for the first $t$ frames with $\bm{G}_0$ being a zero matrix. The weight coefficients before the sum make that each $\bm{C}_p$ for $1 \le p \le t$ contributes equally to $\bm{G}_t$.

\subsubsection{Context Distribution}
At the current frame, our pipeline distributes relevant information stored in the global context module to each pixel. In the distribution process, the global context module first produces the query keys in the same way as the spatce-time memory module. However, the global context module uses the queries differently. At each location, it interprets the query key as a set of weights over the global context features in the memory and computes the weighted sum of the features as the distributed context information. In contrast, the memory read process of the STM module is much more complex. The STM module uses the query to compute a similarity score with the key at each location of each frame in the memory. After that, the module computes the weighted sum of the values in the memory weighted by these similarity scores. Following is an formula that expresses this context distribution process:
\begin{equation}
\label{eq:eq3}
    \bm{D}_t = q(\bm{X}_t)\bm{G}_{t - 1},
\end{equation}
where $\bm{D}_t$ is the distributed global features for frame $t$, and $q$ is the function generating the queries.

Surprisingly, the simple context distribution process of the global context module and the much more complex process of the space-time memory module accomplish the same goal, which is to summarize semantic regions across all past frames that are of interest to the querying location in the current frame. The space-time memory module achieves this by first identifying such regions via query-key matching and then summarizing their values through a weighted sum. The global context module achieves the same much more efficiently since the global context vectors are already global summarizations of regions with similar semantics across all past frames. Therefore, the querying location only needs to determine on an appropriate weighting over the global context vectors to produce a vector that summarizes all the regions of interest to itself. For example, if a pixel is interested in persons, it could place large weights on global context vectors that summarize faces and bodies. Another pixel might be interested in the foreground and could put large weights on vectors that summarize various foreground object categories. We provide a mathematical proof that the global context and space-time memory modules have identical modeling power in the supplementary materials.

\subsubsection{Comparison with the Space-Time Memory Module}

We have plotted the detailed implementation of our global context module in Fig.~\ref{fig:compare} and compare with the space-time memory module used in \cite{stm}. There are a few places our global context has advantages in efficacy over space-time memory.

The global summarization process is the first way the global context module gains efficiency advantage over the space-time memory module. The global context matrix has size $C_M \times C_N$, which tends to be much smaller than the $H \times W \times C_N$
key and $H \times W \times C_M$ value matrices that the space-time memory module produces. The second way the global context module improves efficiency is that it adds the extracted context matrix to the stored context matrix, thereby keeping constant memory usage however many frames are processed. In contrast, the space-time memory module concatenates the obtained key and value matrices to the original key and value matrices in the memory, thus having a linear growth of memory with the number of frames.

For the computation on the current frame, i.e. the context distribution step, our global context module only needs to perform a light weight matrix product of size $C_M \times C_N$ and $C_N \times HW$ with $C_MC_NHW$ multiplications involved. In contrast, the last step of memory read for the STM module calculate a matrix product of size $HWT \times HW$ and $HW \times C_M$ ($C_MH^2W^2T$ multiplications), which has much larger memory usage and computational cost than the global context module and has linear complexities with respect to time $T$. To get a more intuitive comparison of the two, we calculate the computation and resources needed in this step when the input to the encoder is of size $384 \times 384$, $C_M = 512$, and $C_N = 128$ (the default setting in STM). The detailed numbers are listed in Table~\ref{tab:complexity}. It is noticeable that our global context module has great advantages over space-time memory in terms of both computation and memory cost, especially when the number of processed frames $t$ becomes large along the processing.

\begin{table}[tb]
    \centering
    \caption{The complexity comparison of the memory read operation in space-time memory~\cite{stm} and context distribution in our global context module. The memory usage is calculated using the float32 data type.}
    \label{tab:complexity}
     \setlength{\tabcolsep}{6pt}
    \begin{tabular}{crrr|crrrr}
        \toprule
            & $t$ & FLOPS & Memory& & $t$ & FLOPS & Memory\\ 
        \midrule
        STM & $0$ & 0.2 G & 4 MB & GC (ours) & any & \textbf{0.04 G} & \textbf{1 MB} \\
         & $10$ & 2.1 G & 40 MB \\
         & $100$ & 21.2 G & 394 MB \\ 
        \bottomrule
    \end{tabular}
\end{table}

\subsection{Implementation}

\begin{table}[tb]
    \centering
    \caption{Study on the size of global context feature ($C_M \times C_N$). This result is on DAVIS 2016 test set and the $\mathcal{J} \& \mathcal{F}$ is a segmentation accuracy metric (details in Sec. \ref{sec:exp}).}
    \label{tab:ablation}
    \setlength{\tabcolsep}{3pt}
    \begin{tabular}{cccc|cccc}
        \toprule
        $C_M \times C_N$ & $\mathcal{J} \& \mathcal{F}$ & \#Params & Time (s) & $C_M \times C_N$ & $\mathcal{J} \& \mathcal{F}$ & \#Params & Time (s)\\ \midrule
        $512 \times 128$ & \textbf{86.6} & 38 M & 0.040 & $512 \times 512$ & 86.1 & 46 M & 0.046 \\ \bottomrule
    \end{tabular}
\end{table}

Our encoder and decoder design is the same as STM~\cite{stm}. We use ResNet-50~\cite{resnet} as the backbone for both the context encoder and current encoder where the context encoder takes four-channel inputs and the current encoder takes three-channel inputs. The feature map at res4 is used to generate the key and value maps. After the context distribution operation, the features are compressed to 256 channels and fed into the decoder. The decoder takes this low-resolution input feature map and gradually upscales it by a scale factor of two each time. A skip connection is used to retrieve and fuse the feature map at the same resolution in the current encoder with the bilinearly upsampled feature map from the previous stage of the decoder.
The key and value generation (i.e. $k, q, v$ in Equation \eqref{eq:eq1} and \eqref{eq:eq3}) are implemented using $3 \times 3$ convolutions. In our implementation, we set $C_M = 512$ and $C_N = 128$. We have tested with larger feature sizes which introduce more complexities, but we do not observe accuracy gain in segmentation (see Table.~\ref{tab:ablation}).

\subsection{Training}
\label{sec:train}

Our training process mainly contains two stages. We first pre-train the network using simulated videos generated from static images. After that we fine-tune this pre-trained model on the video object segmentation datasets. We minimize the cross-entropy loss and train our model using the Adam~\cite{adam} optimizer with a learning rate of $10^{-5}$.

\subsubsection{Pre-training on images.}
We follow the successful practice in \cite{stm,ranet,rgmp} that pre-trains the network using simulated video clips with frames generated by applying random transformation to static images. We use the images from the MSRA10K~\cite{msra10k}, ECSSD~\cite{ecssd}, and HKU-IS~\cite{hkuis} datasets for the saliency detection task~\cite{borji2015salient}. We found these datasets cover more object categories than those semantic segmentation or instance segmentation datasets~\cite{pascal-voc,sbd,coco}. This is more suitable for our purpose to build a general video object segmentation model. There are in total about 15000 images with object masks. A synthetic clip containing three frames is then generated using image transformations. We use random rotation [$-30^\circ$, $30^\circ$], scaling [$-0.75$, $1.25$], thin plate spline (TPS) warping (as in ~\cite{masktrack}), and random cropping for the video data simulation. We use 4 GPUs and set the batch size to be 8. We run the training for 100 epochs, and it takes one day to finish the pre-training.

\subsubsection{Fine-tuning on videos.}
After training on the synthetic video data, we fine-tune our pre-trained model on video object segmentation datasets~\cite{davis2016,davis2017} at the 480p resolution. The learning rate is set to $10^{-6}$ and we run this fine-tuning for 30 epochs. Each training sample contains several temporally ordered frames sampled from the training video. We use random rotation, scaling, and random frame sampling interval in $[1, 3]$ to gain more robustness to the appearance changes over a long time. We have tested different clip lengths (e.g. 3, 6, 9 frames) but did not observe performance gains on lengths greater than three. Therefore, we stick to three-frame clips. In the training, the network infers the object mask on the second frame and back propagates the error. Then, the soft mask from the network output is fed to the encoder to infer the mask on the third frame without thresholding as in \cite{stm}.

\section{Experimental Results}
\label{sec:exp}

We evaluate our method and compare it with others on the DAVIS~\cite{davis2016,davis2017} and YouTube-VOS~\cite{youtubevos} benchmarks. The object segmentation mask evaluation metrics used in our experiments are the average region similarity ($\mathcal{J}$ mean), the average contour accuracy ($\mathcal{F}$ mean), and the average of the two (${\mathcal{J} \& \mathcal{F}}$ mean). DAVIS 2016~\cite{davis2016} is for single object segmentation. DAVIS 2017~\cite{davis2017} and YouTube-VOS~\cite{youtubevos} contain multiple object segmentation tasks.

\subsection{DAVIS 2016 (Single Object)}

\begin{table}[tb]
    \centering
    \caption{Quantitative comparison on DAVIS 2016 validation set. The results are sorted for online (OL) and non-online methods respectively according to $\mathcal{J} \& \mathcal{F}$ mean. The highest scores in each category are highlighted in bold.}
    \setlength{\tabcolsep}{7pt}
    \begin{tabular}{lccrrr}
        \toprule
        Method & OL & Time (s) & ${\mathcal{J} \& \mathcal{F}}$ & $\mathcal{J}$ Mean & $\mathcal{F}$ Mean\\ \midrule
        OSVOS~\cite{osvos}&\checkmark & 7 & 80.2 & 79.8 & 80.6 \\
        Lucid~\cite{lucid}&\checkmark & - & 83.0 & 83.9 & 82.0 \\
        CINM~\cite{cinm}&\checkmark & $>$30 & 84.2 & 83.4 & 85.0 \\
        OnAVOS~\cite{onavos}&\checkmark & 13 & 85.5 & 86.1 & 84.9 \\
        OSVOS-S~\cite{osvoss}&\checkmark & 4.5 & 86.6 & 85.6 & 87.5 \\
        PReMVOS~\cite{premvos}&\checkmark & $>$30 & \textbf{86.8} & 84.9 & \textbf{88.6} \\
        DyeNet~\cite{dyenet}&\checkmark & 2.32 & - & \textbf{86.2} & - \\ 
        SiamMask~\cite{siammask}& & 0.03 & 70.0 & 71.7 & 67.8 \\
        OSMN~\cite{osmn}& & 0.13 & 73.5 & 74.0 & 72.9 \\
        PML~\cite{pml}& & 0.28 & 77.4 & 75.5 & 79.3 \\
        VidMatch~\cite{videomatch}& & 0.32 & - & 81.0 & - \\
        FAVOS~\cite{fvos}& & 1.8 & 81.0 & 82.4 & 79.5 \\
        FEELVOS~\cite{feelvos}& & 0.5 & 81.7 & 80.3 & 83.1 \\
        RGMP~\cite{rgmp}& & 0.13 & 81.8 & 81.5 & 82.0 \\
        AGAME~\cite{agame}& & 0.07 & 81.9 & 81.5 & 82.2 \\
        RANet~\cite{ranet}& & 0.13 & 85.5 & 85.5 & 85.4 \\
        STM*~\cite{stm}& & 0.15 & 86.5 & 84.8 & \textbf{88.1} \\
        \textbf{GC (ours)}& & 0.04 & \textbf{86.6} & \textbf{87.6} & 85.7\\
        \bottomrule
    \end{tabular}
    \label{tab:davis2016}
\end{table}

DAVIS 2016~\cite{davis2016} is a widely used benchmark dataset for single object segmentation in videos. It contains 50 videos among which 30 videos are for training and 20 are for validation. There are in total 3455 frames annotated with a single object mask for each frame. We use the official split for training and validation.

We list the quantitative results for representative works on DAVIS 2016 validation set in Table~\ref{tab:davis2016}, including the most recent STM~\cite{stm} and RANet~\cite{ranet}. To show their best performance, we directly quote the numbers posted on the benchmark website or in the papers. We can see that online-learning methods can get higher scores in most metrics. However, recent works of STM~\cite{stm} and RANet~\cite{ranet} demonstrate comparable results without online learning. Overall, the scores of our GC are among the top, including getting the highest $\mathcal{J}$ mean score.

\begin{figure}[tb]
    \centering
    \includegraphics[width=0.55\columnwidth]{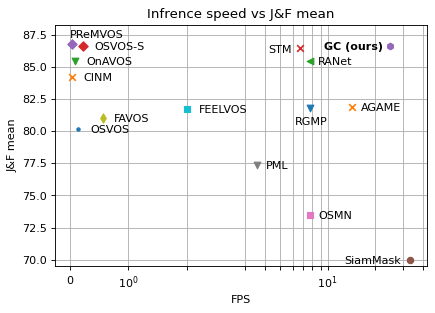}
    \caption{The speed (FPS) \textit{v.s.} accuracy ($\mathcal{J} \& \mathcal{F}$ mean) comparison on DAVIS 2016 validation set at 480p resolution.}
    \label{fig:runtime}
\end{figure}

Furthermore, for a more intuitive comparison, we plot the runtime in terms of average FPS and accuracy in terms of $\mathcal{J} \& \mathcal{F}$ mean for different methods in Fig.~\ref{fig:runtime}. We test the speed on one Tesla P40. It can be seen that although the online-learning methods (\textit{e.g.}~\cite{cinm,premvos,osvoss}) can produce highly accurate results, their speeds are extremely slow due to the time consuming online learning process. The methods without online learning (\textit{e.g.} \cite{pml,rgmp,osmn}) are fast but have lower accuracy. The most recent works STM~\cite{stm} and RANet~\cite{ranet} can get segmentation accuracy comparable to online-learning method while maintaining faster speed (STM~\cite{stm} 6.7 fps, RANet~\cite{ranet} 8.3 fps for 480p frames\footnote{RANet~\cite{ranet} reported a faster runtime in the paper using half-precision computation which is disabled in our test for fair comparison}). Our GC boosts the speed further (25.0 fps) and still maintains high accuracy. Note that the videos in DAVIS datasets are all short ($<$100 frames). If running on longer videos, our speed advantage over STM~\cite{stm} will become more remarkable as STM has a linear time complexity with respect to the video length. While SiamMask~\cite{siammask} is the only method faster than our GC, its accuracy is very unsatisfactory compared to other methods. This demonstrates our GC is both fast and accurate which makes it a practical solution to the video object segmentation problem.

\subsection{DAVIS 2017 (Multiple Objects)}

\begin{table}
    \centering
    \caption{The quantitative comparison on DAVIS 2017 validation set. The results are sorted for online (OL) and non-online methods respectively according to ${\mathcal{J}\&\mathcal{F}}$ Mean. The highest scores in each category are highlighted in bold.}
    \setlength{\tabcolsep}{8 pt}
    \begin{tabular}{lcrrr}
        \toprule
        Method & Online & ${\mathcal{J}\&\mathcal{F}}$ & $\mathcal{J}$ mean & $\mathcal{F}$ mean \\
        \midrule
        OSVOS~\cite{osvos} & \checkmark & 60.3 & 56.6 & 63.9 \\
        OnAVOS~\cite{onavos} & \checkmark & 65.4 & 61.6 & 69.1 \\
        OSVOS-S~\cite{osvoss} & \checkmark  & 68.0  & 64.7  & 71.3 \\
        CINM~\cite{cinm} & \checkmark  & 70.6  & 67.2  & 74.0 \\
        PReMVOS~\cite{premvos} & \checkmark & \textbf{77.8} & \textbf{73.9} &\textbf{81.8} \\
        \addlinespace
        OSMN~\cite{osmn} &  & 54.8  & 52.5  & 57.1 \\
        SiamMask~\cite{siammask} &  & 56.4  & 54.3  & 58.5 \\
        FAVOS~\cite{fvos} &  & 58.2  & 54.6  & 61.8 \\
        VidMatch~\cite{videomatch} & &-  & 56.5 &- \\
        RANet~\cite{ranet} &  & 65.7  & 63.2  & 68.2 \\
        RGMP~\cite{rgmp} &  & 66.7  & 64.8  & 68.6 \\
        FEELVOS~\cite{feelvos} &  & 69.1  & 65.9  & 72.3 \\
        AGAME~\cite{agame} &  & 70.0  & 67.2  & 72.7 \\
        STM*~\cite{stm} & & \textbf{71.6}  & 69.2 & \textbf{74.0} \\
        \textbf{GC (ours)} &  & 71.4 &\textbf{69.3}  & 73.5 \\
        \bottomrule
    \end{tabular}
    \label{tab:davis2017}
\end{table}

DAVIS 2017 is an extension of DAVIS 2016 that contains videos with multiple objects annotated per frame. It has 60 videos for training and 30 videos for testing. We do not use any additional module for multi-object segmentation which \cite{stm,ranet} used, but simply treat each object individually. We still train the network to produce a binary mask for the object. In testing, we use the network to get the soft probability map for each object separately and use a softmax operation as post-processing on the maps for all objects in the frame to produce the multi-label segmentation mask.

Table~\ref{tab:davis2017} summarizes the performance of existing methods and compare them with ours on DAVIS 2017 dataset. The multi-object scenarios is more challenging than the single object ones due to the interactions and occlusions among multiple objects. It can be seen that again online-learning methods, \textit{e.g.}~\cite{cinm,premvos}, get decent scores in all metrics, but have longer runtime. For non-online methods, STM~\cite{stm} ranks the highest overall. Our model can get almost identical performance with STM~\cite{stm} but with faster speed and much less memory consumption.

\subsection{YouTube-VOS}

YouTube-VOS~\cite{youtubevos} is a large-scale dataset for multiple object segmentation in videos. Its training set contains 4453 annotated videos and validation set contains 474 videos. Table~\ref{tab:youtubevos} compares the performance of different methods on this dataset. Note that there are unseen object categories in the validation set. The unseen objects are tested separately to measure the generalization power of each method. It can be seen that STM~\cite{stm} gets remarkable high scores. Our GC is among the top performance tier. Further, the results of our method do not show large performance difference between seen and unseen objects.

\begin{table}[tb]
    \centering
    \caption{The quantitative comparison on YouTube-VOS~\cite{youtubevos} validation set. The results for other methods are quoted from~\cite{agss,stm}.}
    \setlength{\tabcolsep}{1.2pt}
    \begin{tabular}{lrrrrrrrrr}
        \toprule
         & RVOS & OSVOS &S2L(OL) & PreMVOS & AGAME &BoLTVOS & AGSS & STM & \textbf{GC} \\
         & \cite{rvos} &\cite{osvos} &\cite{youtubevos} & \cite{premvos} & \cite{agame} &\cite{boltvos} & \cite{agss} & \cite{stm} & \textbf{(ours)} \\
        \midrule
        Overall & 56.8 & 58.8 & 64.4 & 66.9 & 66.1 & 71.1 & 71.3 & 79.4 & 73.2 \\
        $\mathcal{J}$ seen & 63.6 & 59.8 & 71.0 & 71.4 & 67.8 & 71.6 & 71.3 & 79.7 & 72.6 \\
        $\mathcal{J}$ unseen& 45.5 & 54.2 & 55.5 & 56.5 & 60.8 & 64.3 & 65.5 & 72.8 & 68.9 \\ 
        $\mathcal{F}$ seen & 67.2 & 60.5 & 70.0 & 75.9 & - & - & 76.2 & 84.2 & 75.6 \\
        $\mathcal{F}$ unseen& 51.0 & 60.7 & 61.2 & 63.7 & - & - & 73.1 & 80.9 & 75.7 \\
        \bottomrule
    \end{tabular}
    \label{tab:youtubevos}
\end{table}

Notably, the performance gap between STM and our GC does not come from the models. Instead, two external factors are the main cause of the gap:
\begin{itemize}
    \item an easier testing protocol used by STM;
    \item a soft-aggregation post-processing module, which is compatible with GC but not implemented due to time constraints.
\end{itemize}
Table \ref{tab:vs-stm} summarized the comparison.

\begin{table}[]
    \centering
    \caption{Cause of performance gap between STM and our GC on YouTube-VOS.}
    \label{tab:vs-stm}
    \begin{tabular}{l@{\hskip 12pt}c@{\hskip 12pt}r@{\hskip 12pt}r}
        \toprule
        Model & Soft aggregation & Test input stride& $\mathcal{J}\&\mathcal{F}$ mean \\
        \midrule
        GC & & 5 & 73.2 \\
        STM & & 5 & 73.7 \\
        STM & \checkmark & 1 & 79.4 \\
        \bottomrule
    \end{tabular}
\end{table}

\subsection{Qualitative Results}

\begin{figure*}[tb]
    \centering
    \setlength{\tabcolsep}{1 pt}
    \begin{tabular}{cccccc}
        \includegraphics[width=0.16\linewidth]{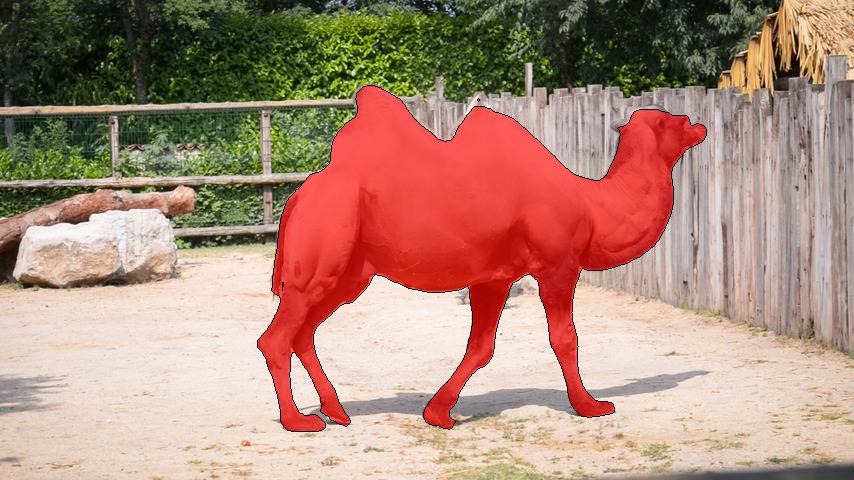} &
        \includegraphics[width=0.16\linewidth]{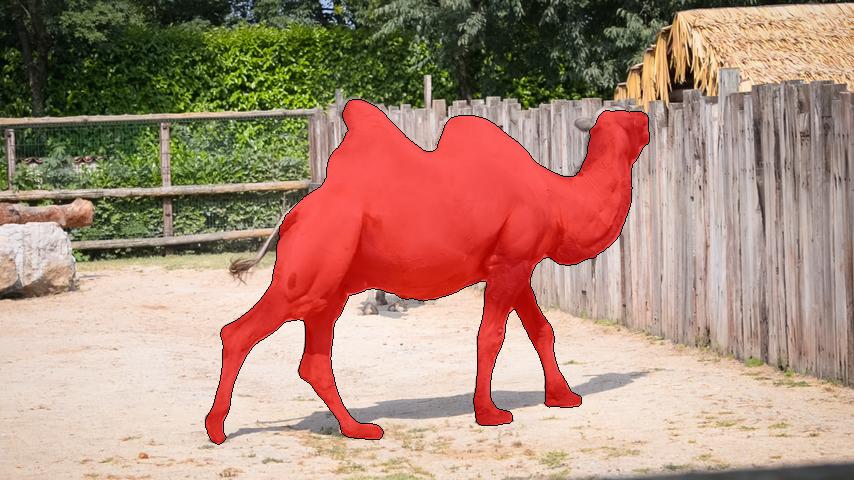} &
        \includegraphics[width=0.16\linewidth]{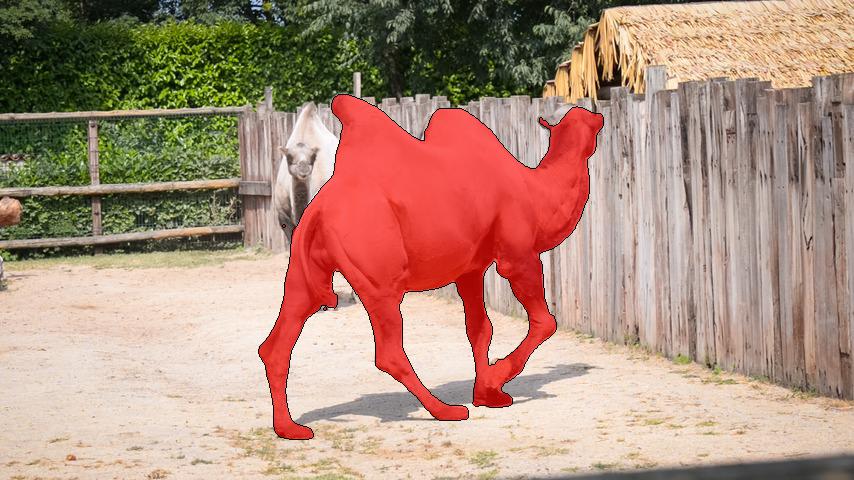} &
        \includegraphics[width=0.16\linewidth]{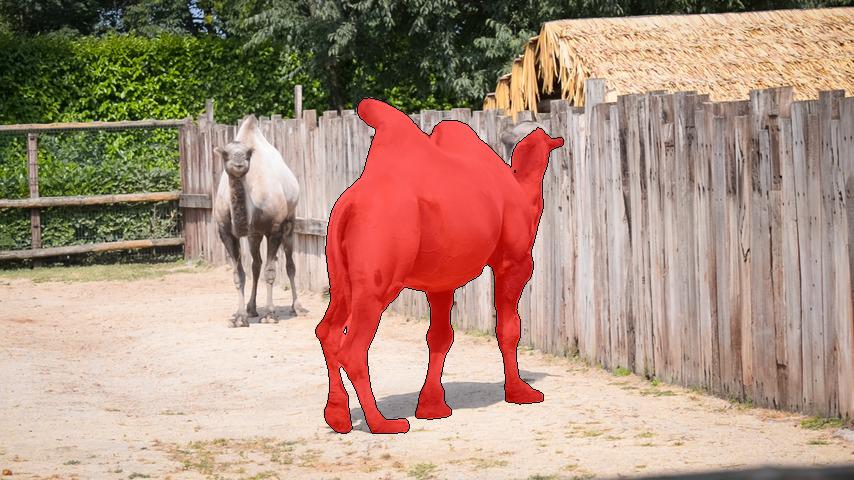} &
        \includegraphics[width=0.16\linewidth]{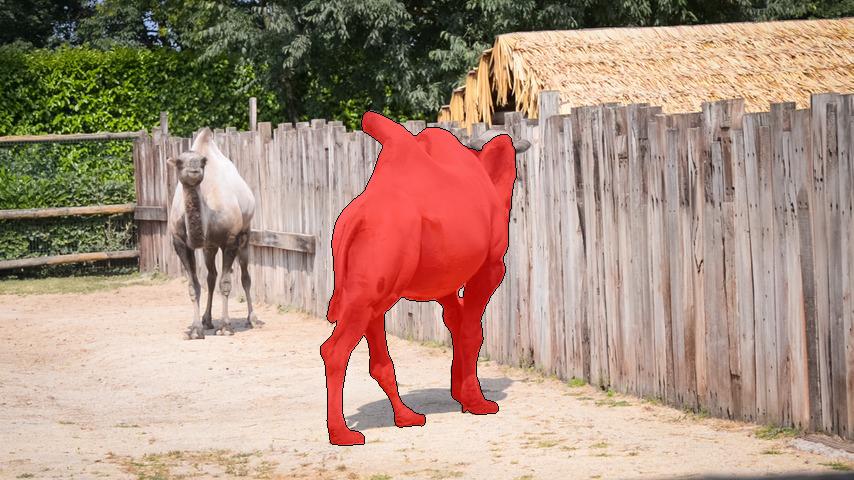} &
        \includegraphics[width=0.16\linewidth]{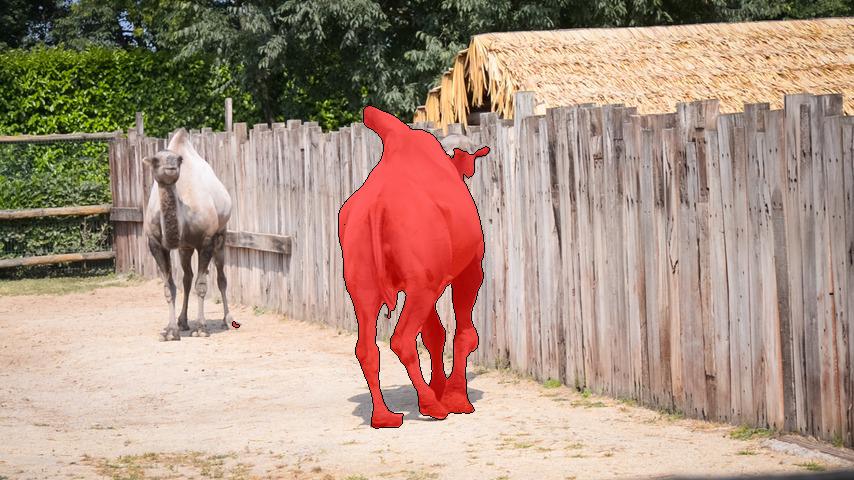} \\
        \includegraphics[width=0.16\linewidth]{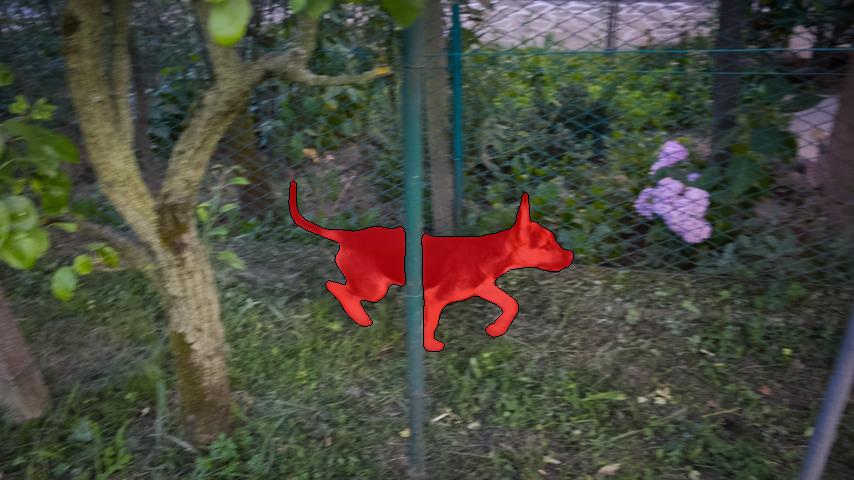} &
        \includegraphics[width=0.16\linewidth]{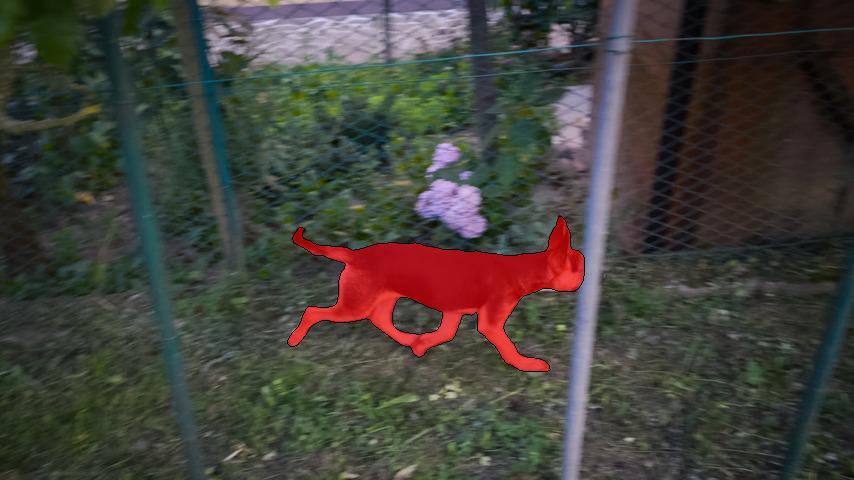} &
        \includegraphics[width=0.16\linewidth]{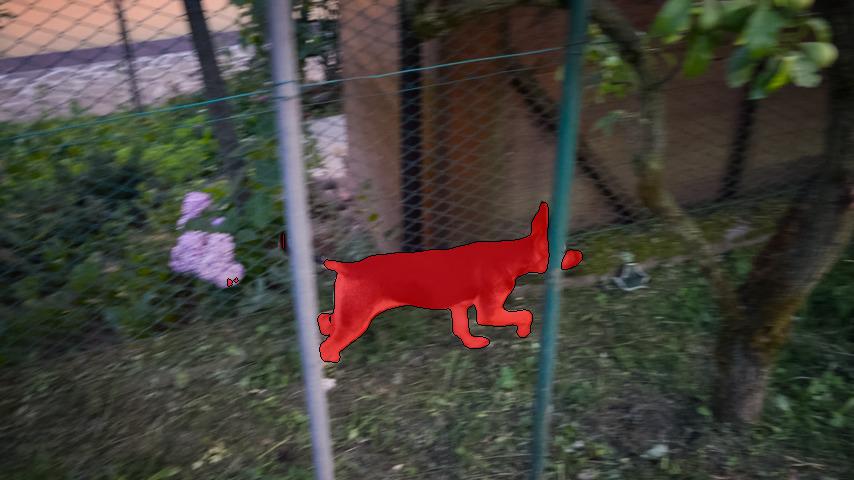} &
        \includegraphics[width=0.16\linewidth]{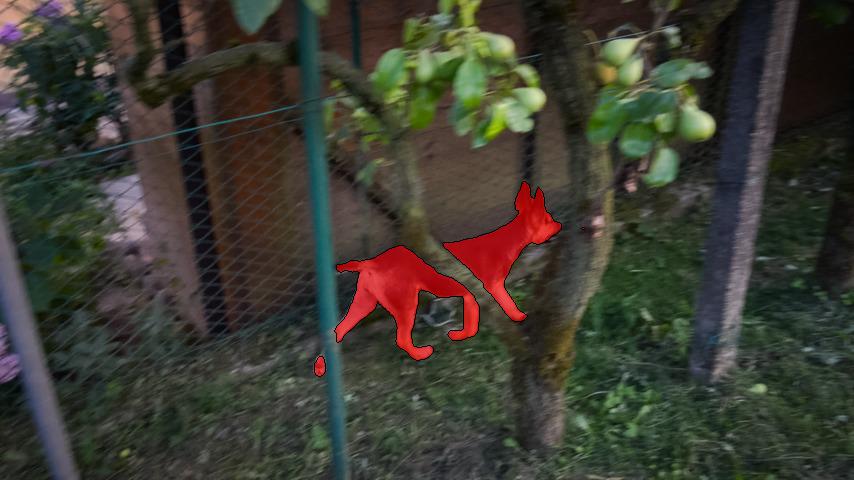} &
        \includegraphics[width=0.16\linewidth]{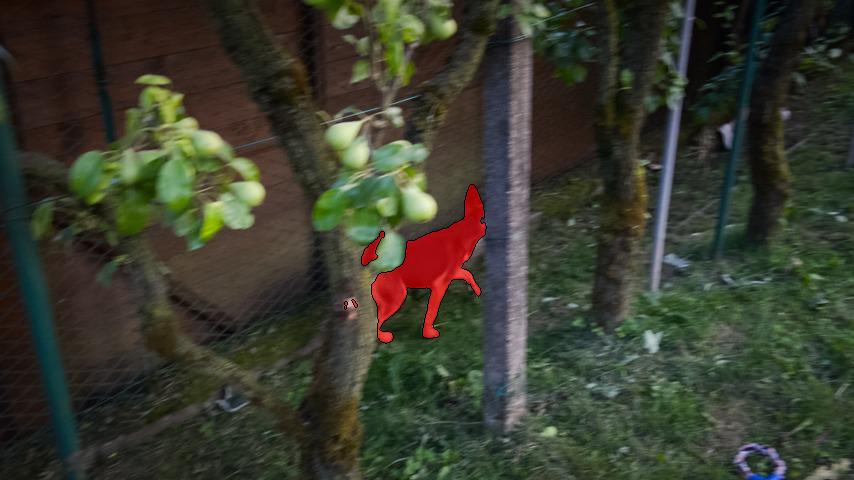} &
        \includegraphics[width=0.16\linewidth]{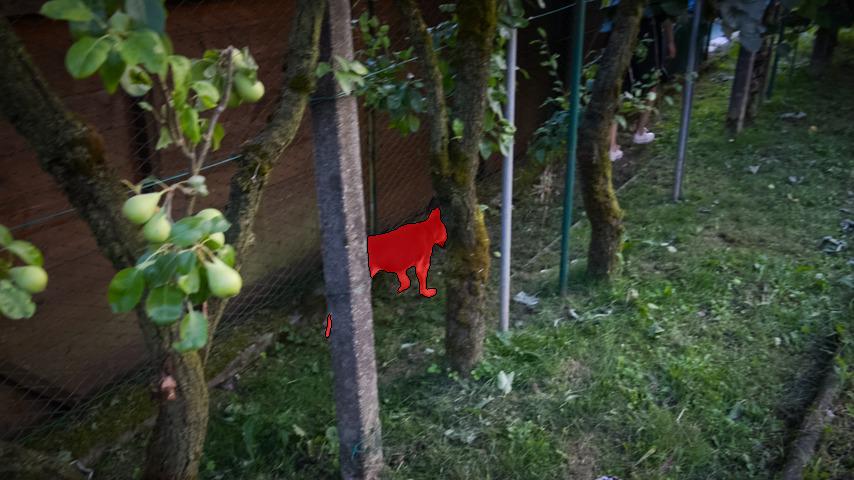} \\
        \includegraphics[width=0.16\linewidth]{vis/drift-chicane/00001.jpg} &
        \includegraphics[width=0.16\linewidth]{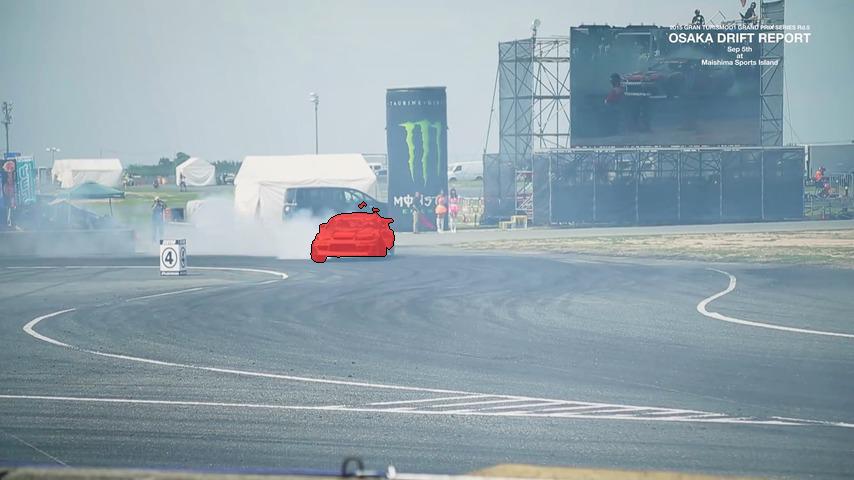} &
        \includegraphics[width=0.16\linewidth]{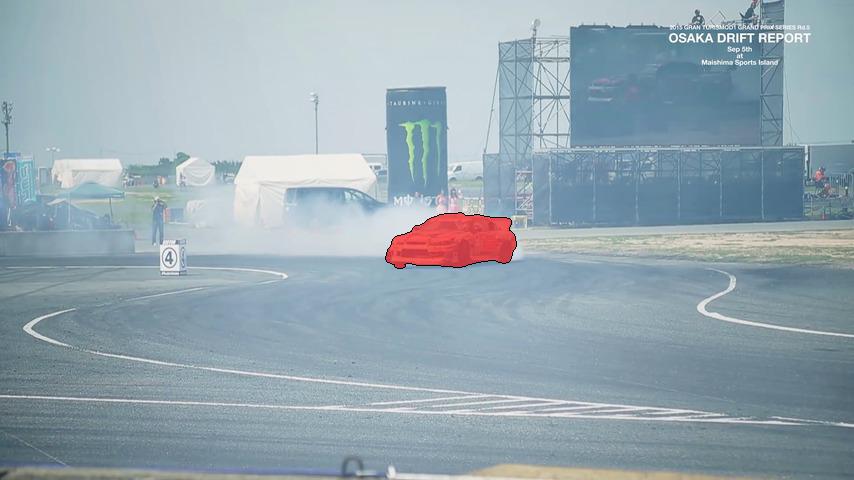} &
        \includegraphics[width=0.16\linewidth]{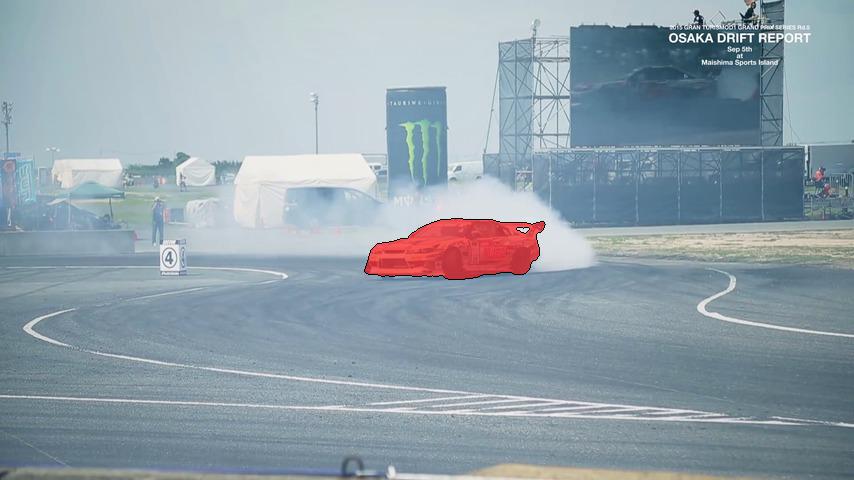} &
        \includegraphics[width=0.16\linewidth]{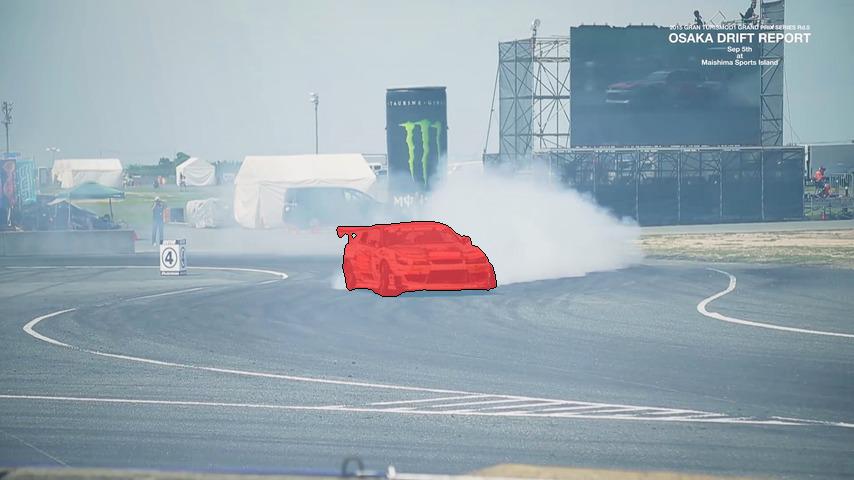} &
        \includegraphics[width=0.16\linewidth]{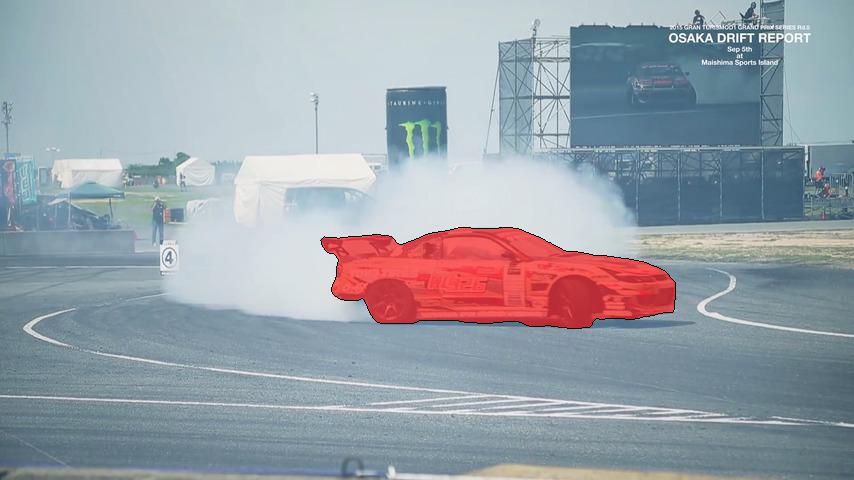} \\
        \includegraphics[width=0.16\linewidth]{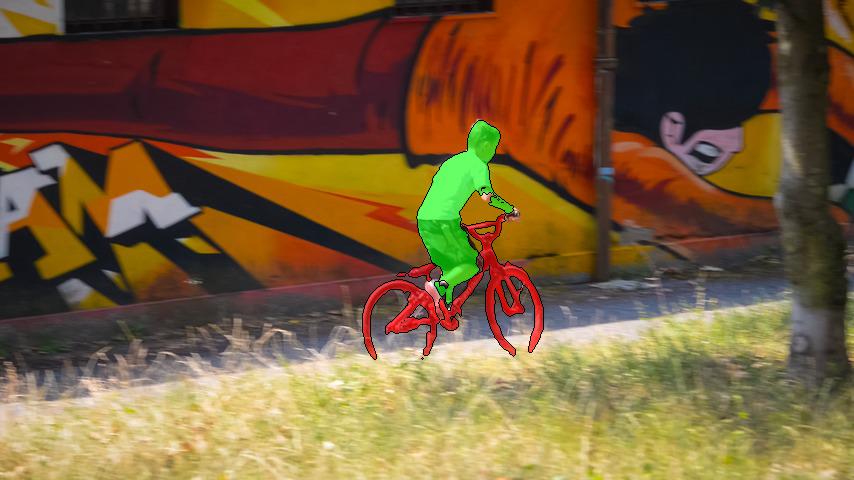} &
        \includegraphics[width=0.16\linewidth]{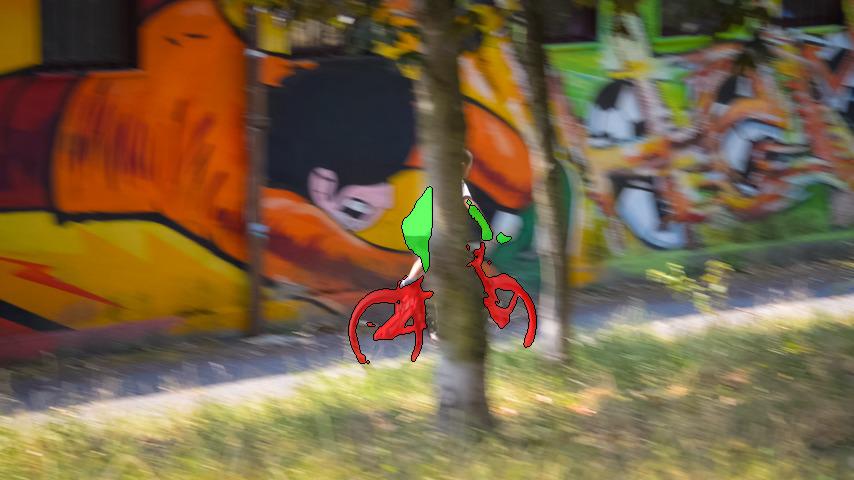} &
        \includegraphics[width=0.16\linewidth]{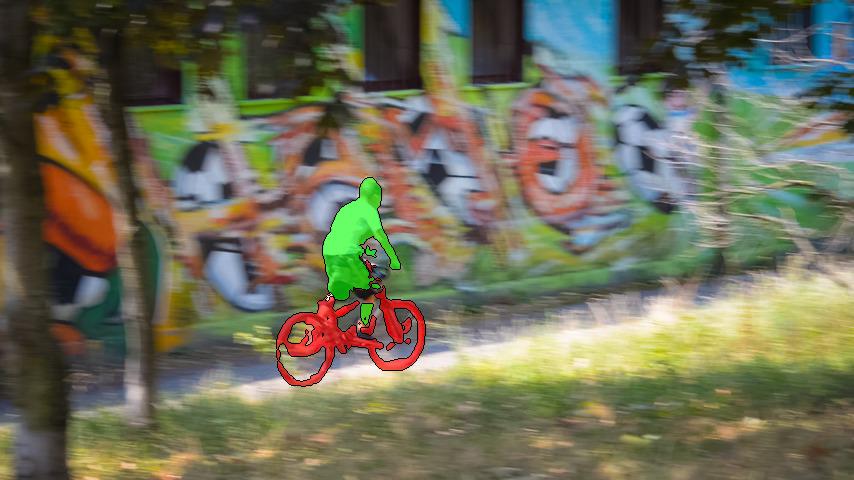} &
        \includegraphics[width=0.16\linewidth]{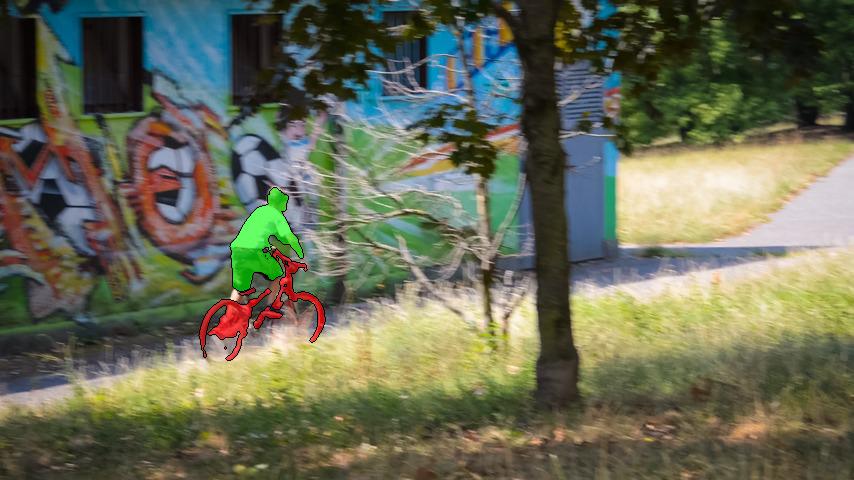} &
        \includegraphics[width=0.16\linewidth]{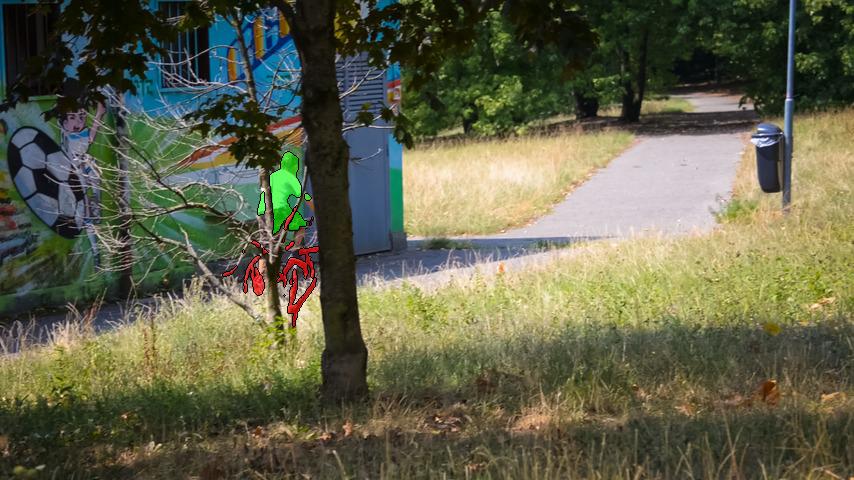} &
        \includegraphics[width=0.16\linewidth]{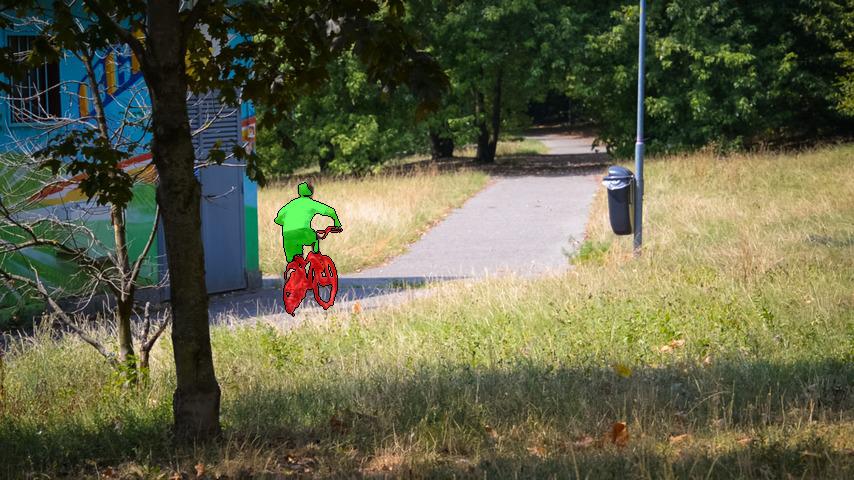} \\
        \includegraphics[width=0.16\linewidth]{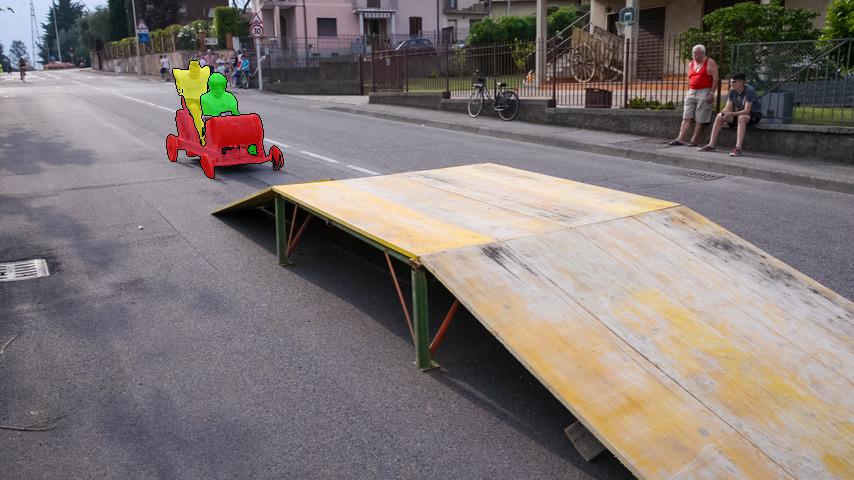} &
        \includegraphics[width=0.16\linewidth]{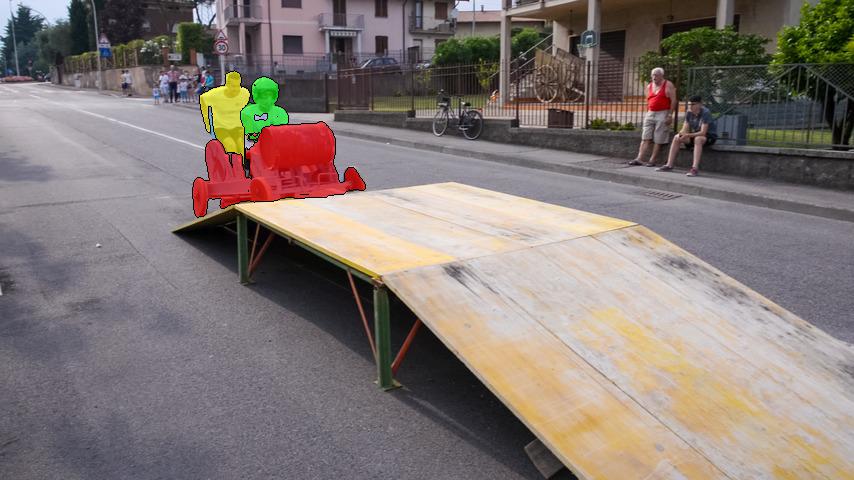} &
        \includegraphics[width=0.16\linewidth]{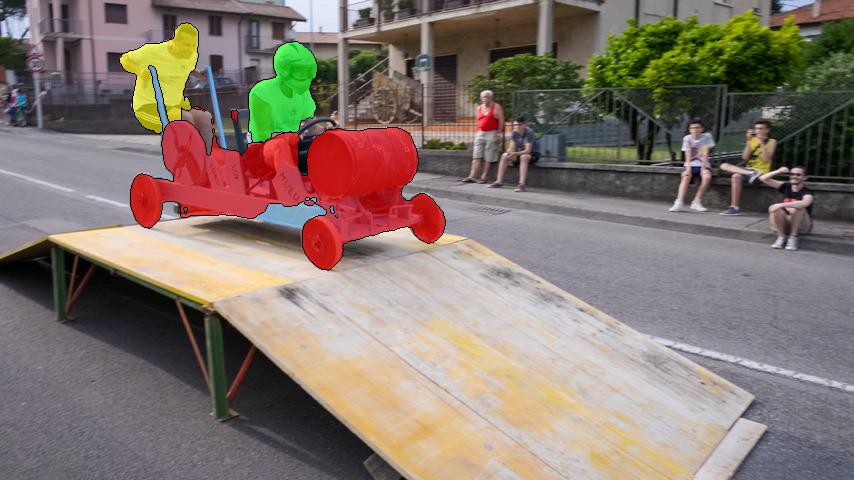} &
        \includegraphics[width=0.16\linewidth]{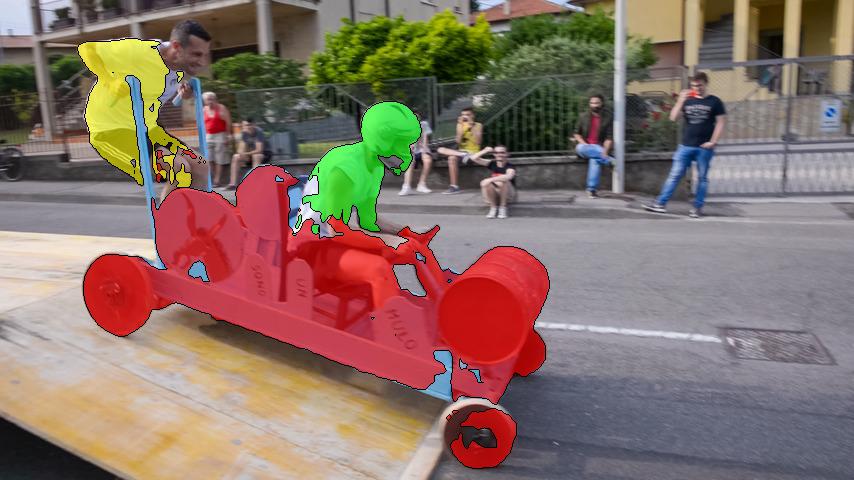} &
        \includegraphics[width=0.16\linewidth]{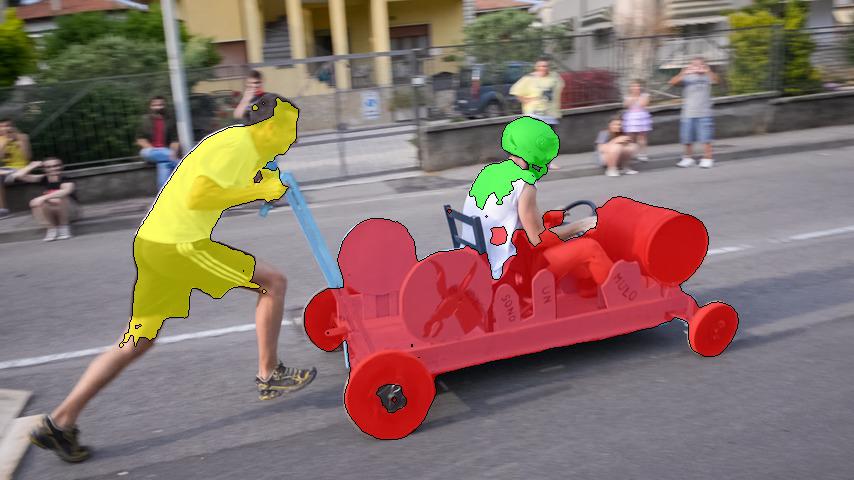} &
        \includegraphics[width=0.16\linewidth]{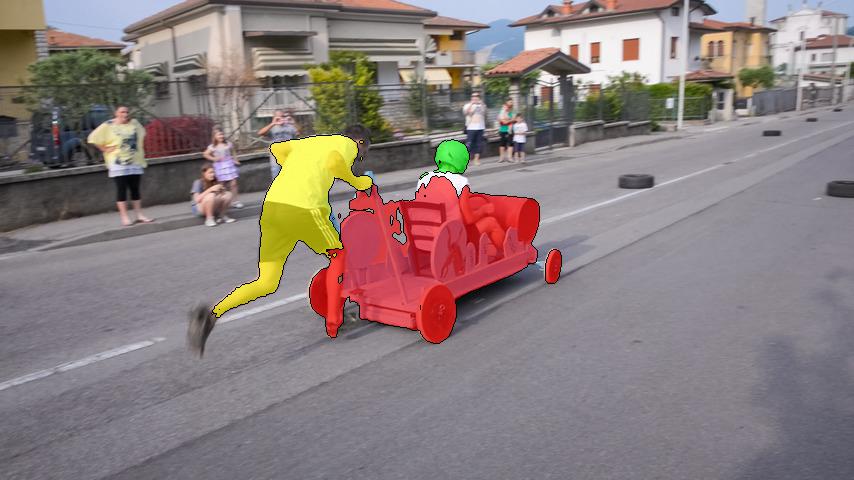} \\
        \includegraphics[width=0.16\linewidth]{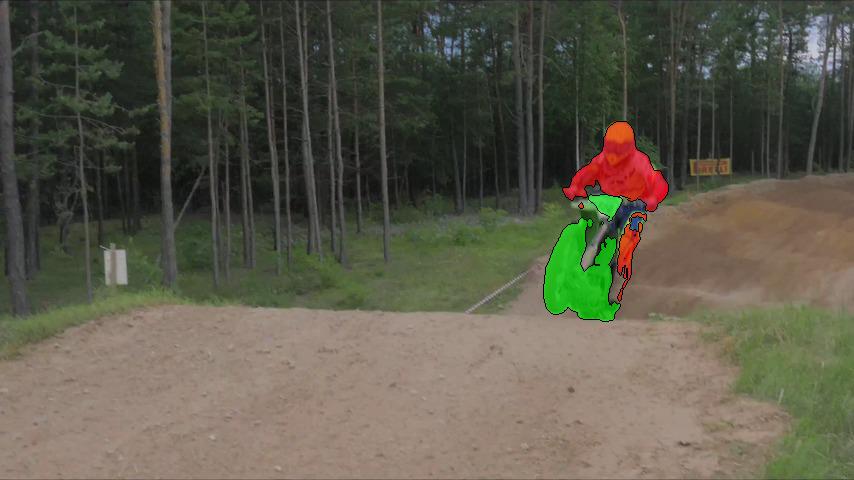} &
        \includegraphics[width=0.16\linewidth]{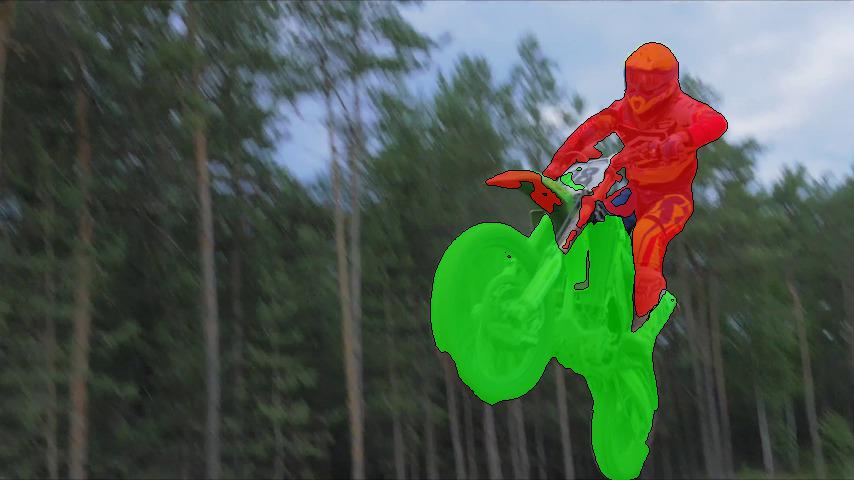} &
        \includegraphics[width=0.16\linewidth]{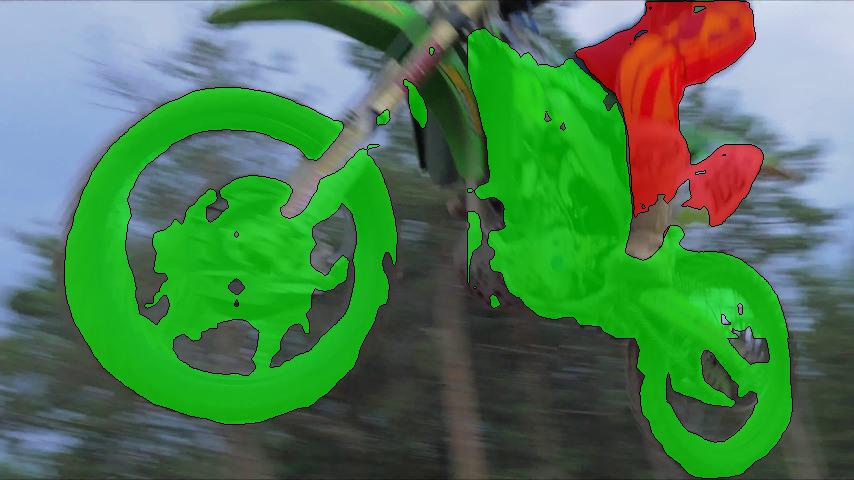} &
        \includegraphics[width=0.16\linewidth]{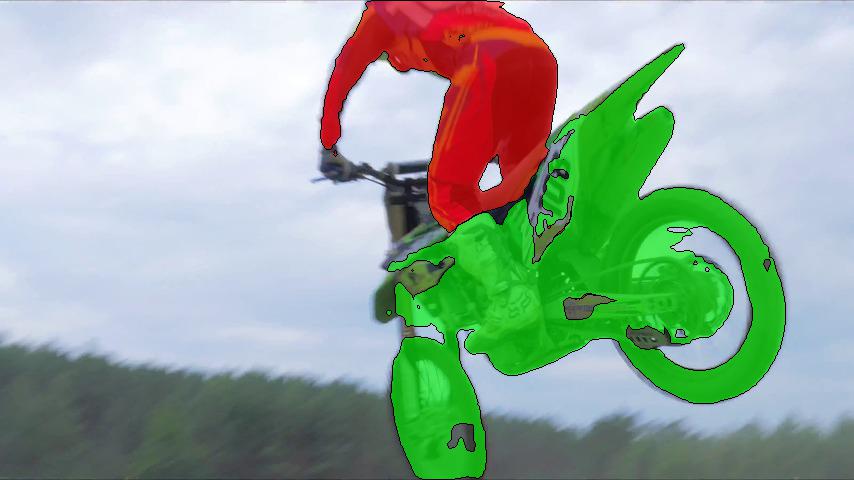} &
        \includegraphics[width=0.16\linewidth]{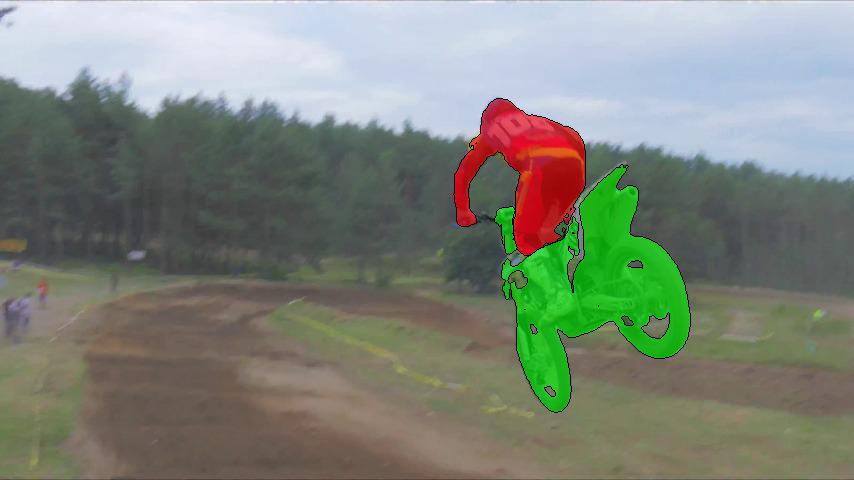} &
        \includegraphics[width=0.16\linewidth]{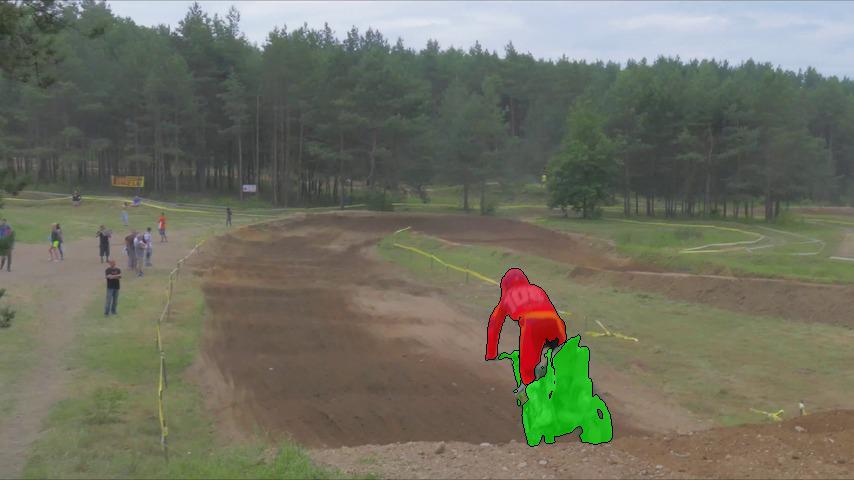}
    \end{tabular}
    \caption{The visual results of video object segmentation using our global context module.}
    \label{fig:davis}
\end{figure*}

Fig.~\ref{fig:davis} shows visual examples of our segmentation results. As can be seen in the figure, our global context module can effectively handle many challenging cases such as appearance changes (row 1), size changes (row 3, 5, and 6), and occlusions (row 2 and 4).

\begin{figure}[tb]
    \centering
    \setlength{\tabcolsep}{1 pt}
    \begin{tabular}{ccccc}
        \includegraphics[width=0.1945\linewidth, height=42 pt]{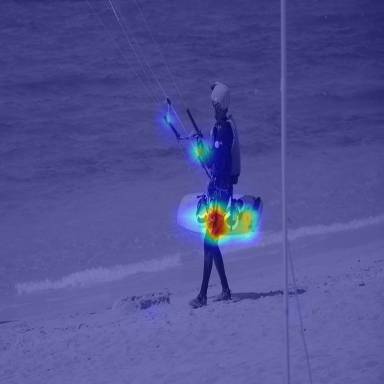} &
        \includegraphics[width=0.1945\linewidth, height=42 pt]{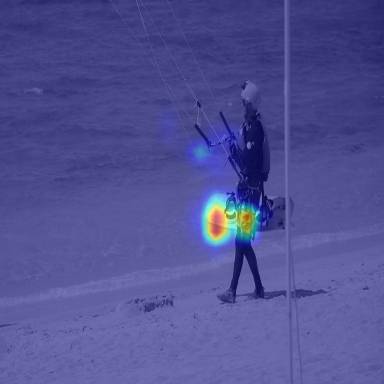} &
        \includegraphics[width=0.1945\linewidth, height=42 pt]{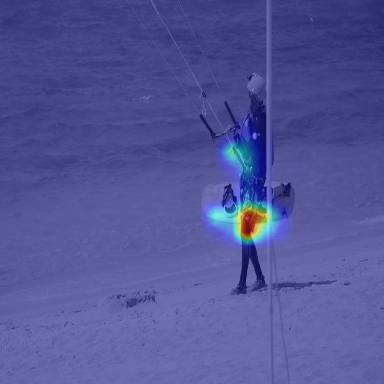} &
        \includegraphics[width=0.1945\linewidth, height=42 pt]{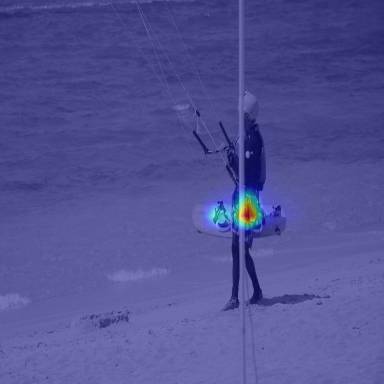} &
        \includegraphics[width=0.1945\linewidth, height=42 pt]{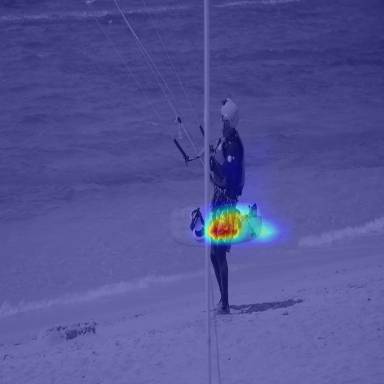} \\
        \includegraphics[width=0.1945\linewidth, height=42 pt]{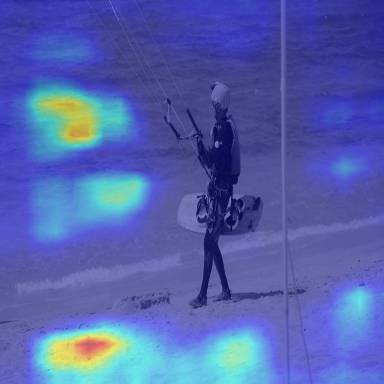} &
        \includegraphics[width=0.1945\linewidth, height=42 pt]{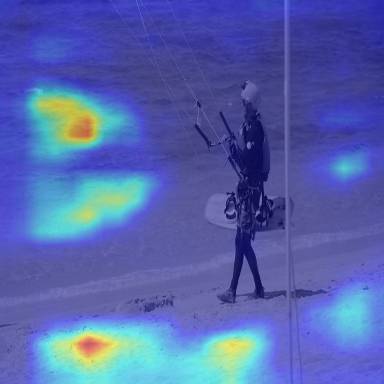} &
        \includegraphics[width=0.1945\linewidth, height=42 pt]{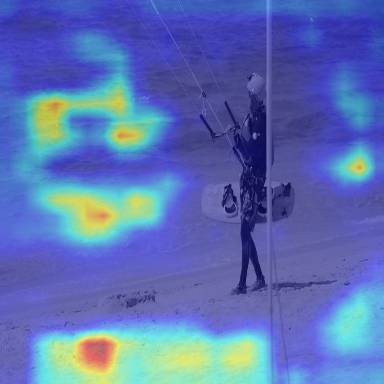} &
        \includegraphics[width=0.1945\linewidth, height=42 pt]{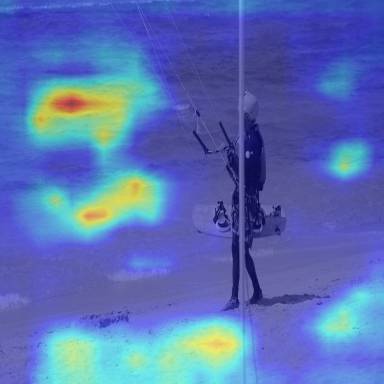} &
        \includegraphics[width=0.1945\linewidth, height=42 pt]{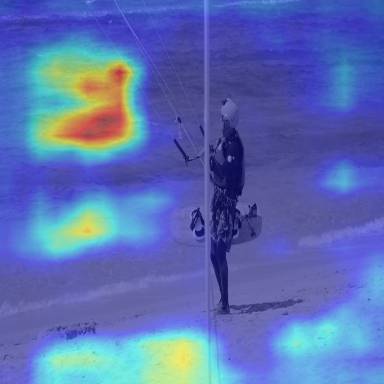} \\
    \end{tabular}
    \caption{Visualization of the global context keys.}
    \label{fig:vis}
\end{figure}

\subsection{Visualization of the Global Context Module}

Fig.~\ref{fig:vis} plots visualization of the global context module. As described in Sec. 3.1, each channel of the global context key (the keys in the blue region in Fig.~\ref{fig:compare} left) is an attention map (or weight map) over all spatial locations. Our global context module aggregates the features at these locations to form one global context feature vector by a weighted sum. After such aggregation on all $\bm{C}_N$ channels, $\bm{C}_N$ feature vectors are generated. Fig.~\ref{fig:vis} shows the visualization of two channels in the global context key at evenly sampled time in a whole video sequence. We can clearly see that it can summarize the segmentation information well, where the keys in the upper row capture parts of the foreground object and the keys in the bottom capture the background. In addition, it shows that the global context module can capture the foreground and background information consistently throughout the video. 

\section{Conclusion}
\label{sec:conclusion}

We have presented a practical solution to the problem of semi-supervised video object segmentation. It is achieved by a novel global context module that effectively and efficiently captures the object segmentation information in all processed frames with a set of fixed-size features. The evaluation on multiple benchmark datasets shows that our method gets top performance, especially on the single object DAVIS 2016 dataset, and runs at a much faster speed than all top-performing methods, including the state-of-the-art STM. Our global context module is also efficient in memory usage and will not have memory issues as with STM for longer video sequences. We believe that our global context module has the potential to become a core module in practical video object segmentation tools. In the future, we want to optimize it further to make it suitable for running on portable devices like tablets and mobile phones. Applying the global context module to other video-related computer vision problems is also of our interest.

\clearpage

\bibliographystyle{splncs04}
\bibliography{egbib}

\clearpage
\appendix

\section{Proof of Equivalence between the Global Context Module and the Space-Time Memory Module}
\label{sec:method}

\subsection{Mathematical Formulations of the Global Context Module and the Space-Time Memory Module}

The global context (GC) module consists of three steps, namely context extraction, context update, and context distribution. The context extraction step can be expressed as
\begin{equation}
\label{eq:feat-prod}
    \bm{C}_t = k(\bm{X}_t)^\mathsf{T}v(\bm{X}_t),
\end{equation}
where $t$ is the index of the current frame, $\bm{X}_t$ is the input feature, $\bm{C}_t$ is the global context feature of the current frame, and $k, v$ are the key and value generation functions. The context update step is
\begin{equation}
\label{eq:feat-update}
    \bm{G}_t = \frac{t - 1}{t}\bm{G}_{t - 1} + \frac{1}{t}\bm{C}_t,
\end{equation}
where $\bm{G}_t$ is the global context feature of the first $t$ frames. The context distribution step is
\begin{equation}
\label{eq:feat-retriev}
    \bm{R}_t = q(\bm{X}_t)\bm{G}_{t - 1},
\end{equation}
where $\bm{R}_t$ is the distributed global context feature and $q$ is the query generation function.

The space-time memory (STM)~\cite{stm} module similarly consists of three steps, namely memory production, memory write, and memory read. The memory production step can be expressed as
\begin{align}
\begin{split}
\label{eq:mem-prod}
    \bm{K}_{P, t} &= k(\bm{X}_t),\\
    \bm{V}_{P, t} &= v(\bm{X}_t),
\end{split}
\end{align}
where $t$ is the index of the current frame, $\bm{X}_t$ is the input feature, $\bm{K}_{M, t}, \bm{V}_{M, t}$ are the key and value produced from the current frame, and $k, v$ are the key and value generation functions. The memory write operation is
\begin{align}
\begin{split}
\label{eq:mem-write}
    \bm{K}_{M, t} &= \bm{K}_{M, t - 1} \odot \bm{K}_{P, t},\\
    \bm{V}_{M, t} &= \bm{V}_{M, t - 1} \odot \bm{V}_{P, t},
\end{split}
\end{align}
where $\bm{K}_{M, t}, \bm{V}_{M, t}$ are the key and value in the memory from the first $t$ frames and $\odot$ denotes concatenation along the spatial dimension. The memory read step is
\begin{equation}
\label{eq:mem-read}
    \bm{E}_t = \frac{1}{t}\left(q(\bm{X}_t)\bm{K}_{M, t - 1}^\mathsf{T}\right)\bm{V}_{M, t - 1},
\end{equation}
where $\bm{E}_t$ is the memory reading and $q$ is the query generation function.

\subsection{Proof of Equivalence}

Expanding Equation \eqref{eq:feat-update} by substituting it to itself gives
\begin{equation}
\label{eq:feat-avg}
    \bm{G}_{t - 1} = \frac{1}{t - 1}\sum_{f = 1}^{t - 1}{\bm{C}_f}.
\end{equation}
Note that $\bm{G}_0$ is the zero matrix. Substituting Equation \eqref{eq:feat-avg} into Equation \eqref{eq:feat-retriev} results in
\begin{equation}
\label{eq:feat-retriev-avg}
    \bm{R}_t = \frac{1}{t - 1}q(\bm{X}_t)\sum_{f = 1}^{t - 1}{\bm{C}_f}.
\end{equation}
Combining Equations \eqref{eq:feat-prod} and \eqref{eq:feat-retriev-avg}, we have
\begin{equation}
\label{eq:feat-all}
    \bm{R}_t = \frac{1}{t - 1}q(\bm{X}_t)\sum_{f = 1}^{t - 1}{k(\bm{X}_f)^\mathsf{T}v(\bm{X}_f)}.
\end{equation}

Similarly, expanding Equations \eqref{eq:mem-write} gives
\begin{align}
\begin{split}
\label{eq:mem-concat}
    \bm{K}_{M, t - 1} &= \bigodot_{f = 1}^{t - 1}\bm{K}_{P, f},\\
    \bm{V}_{M, t - 1} &= \bigodot_{f = 1}^{t - 1}\bm{V}_{P, f},
\end{split}
\end{align}
where $\bigodot$ stands for concatenation along the spatial dimension. Substituting Equations \eqref{eq:mem-prod} into Equation \eqref{eq:mem-concat} results in
\begin{align}
\begin{split}
\label{eq:mem-concat1}
    \bm{K}_{M, t - 1} &= \bigodot_{f = 1}^{t - 1}k(\bm{X}_f),\\
    \bm{V}_{M, t - 1} &= \bigodot_{f = 1}^{t - 1}v(\bm{X}_f).
\end{split}
\end{align}
Combining Equations \eqref{eq:mem-read} and \eqref{eq:mem-concat1}, we have
\begin{align}
\begin{split}
\label{eq:mem-all}
    \bm{E}_t &= \frac{1}{t - 1}\left(q(\bm{X}_t)\bigodot_{f = 1}^{t - 1}k(\bm{X}_f)^\mathsf{T}\right)\bigodot_{f = 1}^{t - 1}v(\bm{X}_f)\\
    &= \frac{1}{t - 1}\left(\bigodot_{f = 1}^{t - 1}q(\bm{X}_t)k(\bm{X}_f)^\mathsf{T}\right)\bigodot_{f = 1}^{t - 1}v(\bm{X}_f)\\
    &= \frac{1}{t - 1}\sum_{f = 1}^{t - 1}q(\bm{X}_t)k(\bm{X}_f)^\mathsf{T}v(\bm{X}_f)\\
    &= \frac{1}{t - 1}q(\bm{X}_t)\sum_{f = 1}^{t - 1}k(\bm{X}_f)^\mathsf{T}v(\bm{X}_f)
\end{split}
\end{align}
Comparing Equations \eqref{eq:feat-all} and \eqref{eq:mem-all}, we have
\begin{equation}
\label{eq:qed}
    \bm{R}_t = \bm{E}_t,
\end{equation}
which completes the proof.

\subsection{Impact of Softmax Normalization}

In practice, in addition to the operations in \eqref{eq:mem-read}, the STM module usually uses a softmax along the rows on $q(\bm{X}_t)\bm{K}^\mathsf{T}_{M, t - 1}$. This operation makes each row sum up to 1. The interpretation of this normalization is that it makes each row, which corresponds to a query pixel, represent a probabilistic distribution of attention over all locations in the memory.

Since softmax is non-linear, the GC module cannot perfectly recreate its effect. However, the GC module applies two softmax operations, one on each row of $q(\bm{X}_t)$ and the other on each column of $k(\bm{X}_t)$. Then, if one hypothetically multiplies these two normalized matrices, the product matrix will share the critical property of $\operatorname{softmax}(q(\bm{X}_t)\bm{K}^\mathsf{T}_{M, t - 1})$, i.e. each row sums up to 1 and represents a probabilistic distribution of attention for the query pixel over all locations in the past frames. Therefore, the two operations together are an effective approximation of the single softmax in STM. Thus, the GC module remains approximately equivalent to the STM module even with the presence of the softmax operations. 

\end{document}